\documentclass{IMAIAI}
\usepackage[square,numbers]{natbib}

\usepackage{amsthm}
\usepackage{amsmath}
\usepackage{amssymb}
\usepackage{bm}
\usepackage{mathrsfs}
\usepackage[dvips]{graphicx}

\usepackage{algorithmic}
\usepackage{algorithm}

\usepackage{color}

\usepackage{url}

\usepackage{supertabular}

\newtheorem{thm}{Theorem}

\theoremstyle{definition}

\theoremstyle{remark}

\numberwithin{thm}{section}

\DeclareMathAlphabet{\mathsfsl}{OT1}{cmss}{m}{sl}

\renewcommand{\phi}{\varphi}

\newcommand{\eps}{\varepsilon}

\newcommand{\argmin}{\operatorname*{arg\; min}}

\newcommand{\mtx}[1]{\bm{#1}}

\newcommand{\trace}{\operatorname{tr}}

\newcommand{\Proj}{\ensuremath{\mtx{\Pi}}}

\newcommand{\dist}{\operatorname{dist}}

\newcommand{\bx}{\boldsymbol{x}}

\newcommand{\bP}{\boldsymbol{P}}
\newcommand{\bQ}{\boldsymbol{Q}}

\newcommand{\sX}{\mathcal{X}}
\newcommand{\sI}{\mathcal{I}}

\newcommand{\bSigma}{\boldsymbol\Sigma}

\newcommand{\bU}{\boldsymbol{U}}

\def\reals{\mathbb{R}}
\def\bx{\boldsymbol{x}}
\def\b0{\mathbf{0}}
\def\bP{\boldsymbol{P}}
\def\bQ{\boldsymbol{Q}}
\def\bSigma{\boldsymbol\Sigma}

\def\bU{\boldsymbol{U}}

\def\bv{\boldsymbol{v}}
\def\bA{\boldsymbol{A}}
\def\bT{\boldsymbol{T}}

\def\bI{\mathbf{I}}
\def\b1{\mathbf{1}}
\def\b0{\mathbf{0}}

\def\im{\mathrm{im}}
\def\tr{\mathrm{tr}}

\def\SPSD{\mathrm{S_+}}
\def\PSD{\mathrm{S_{++}}}

\usepackage{algorithmic}
\usepackage{algorithm}
\newcommand{\di}{{\,\mathrm{d}}}
\usepackage{setspace}
\begin{document}
%
\title{Robust subspace recovery by Tyler's M-estimator}

\shorttitle{Robust subspace recovery by Tyler's M-estimator} 
\shortauthorlist{Teng Zhang} 

\author{{
\sc Teng Zhang}$^*$,\\[2pt]
Department of Mathematics, University of Central Florida,
Orlando, Florida 32816, USA\\
$^*${\email{Corresponding author: teng.zhang@ucf.edu}}}

\maketitle

\begin{abstract}
{This paper considers the problem of robust subspace recovery: given a set of $N$ points in $\reals^D$, if many lie in a $d$-dimensional subspace, then can we recover the underlying subspace? We show that Tyler's M-estimator can be used to recover the underlying subspace, if the percentage of the inliers is larger than $d/D$ and the data points lie in general position. Empirically, Tyler's M-estimator compares favorably with other convex subspace recovery algorithms in both simulations and experiments on real data sets.}
{M-estimator, subspace recovery, robust statistics}
\\
2000 Math Subject Classification: 62-07, 62H12,  90C25
\end{abstract}


%

\section{Introduction}
A fundamental problem in data analysis is to approximate a given data set by a subspace, i.e., subspace recovery. The standard approach for subspace recovery is Principal Component Analysis (PCA). However, PCA is problematic when the given data set is corrupted with outliers. Therefore, it is important to develop subspace recovery methods that are robust to outliers, and the purpose of this work is to show that Tyler's M-estimator has theoretical guarantees on robust subspace recovery and performs well empirically.

\subsection{Notation and conventions}
Let $\sX=\{\bx_i\}_{i=1}^N\subset\reals^D$ be a set of $N$ points, and for technical reasons, we assume that $\sX$ does not contain the origin. 
For a $d$-dimensional subspace $L$, we define the projector matrix $\Proj_L$ as the $D\times D$ symmetric matrix such that $\Proj_L^2=\Proj_L$, and the range of $\Proj_L$ is $L$. We define $\bP_{L}$ as any $D\times d$ projection matrix such that  $\Proj_L=\bP_L\bP_L^T$. While $\bP_{L}$ is not uniquely defined, different choices of $\bP_{L}$ will not affect the results in the rest of the paper. We use $L^\perp$ to denote the orthogonal subspace of $L$.

We use $\sX\cap L$ to express the set of points that lie both in $\sX$ and  the subspace $L$, and $\sX\setminus L$ to express the set of points that lie in $\sX$ but not in the subspace $L$.  We use $|\sX|$ to denote the cardinality of the set $\sX$, and $\SPSD(D)$, $\PSD(D)$ to denote the set of $D\times D$ semi-positive definite matrices and the set of $D\times D$ positive definite matrices.
\subsection{Tyler's M-estimator}
Tyler's M-estimator~\cite{Tyler1987} is defined by
\begin{align}\label{eq:problem}
&\bSigma_*=\argmin_{\trace(\bSigma)=1,\bSigma=\bSigma^T,\bSigma\in\PSD(D)}F(\bSigma),\,\,\,\text{where}\,\,\\&F(\bSigma)=\frac{1}{N}\sum_{\bx\in\sX} \log(\bx^T\bSigma^{-1}\bx) + \frac{1}{D}\log\det(\bSigma),\nonumber
\end{align}
and~\cite{Tyler1987} also gives the following iterative algorithm:
\begin{equation}\label{eq:IRLS}
\bSigma^{(k+1)}=\sum_{\bx\in\sX}\frac{\bx\bx^T}{\bx^T\bSigma^{(k)\,-1}\bx}/\trace\Big(\sum_{\bx\in\sX}\frac{\bx\bx^T}{\bx^T\bSigma^{(k)\,-1}\bx}\Big).
\end{equation}

Historically, M-estimators are viewed
as being a more general class than the MLE estimators, and M-estimators of covariance~\cite{Maronna1976, huber_book, robust_stat_book2006} are motivated from the MLE estimators under the assumption that data samples are i.i.d. drawn from the elliptical distribution $C(\rho)e^{-\rho(\bx ^T\bSigma^{-1}\bx)}/\sqrt{\det(\bSigma)},$
where $C(\rho)$ is a normalization constant. That is, M-estimator of covariance is defined as the minimizer of
 \begin{equation}
\frac{1}{N}\sum_{\bx\in\sX}\rho(\bx^T\bSigma^{-1}\bx)+\frac{1}{2}\log\det(\bSigma).
\end{equation}


%
%

%
%
Tyler's M-estimator is a special case of the M-estimators of covariance with $\rho(x)=\frac{D}{2}\log(x)$, and can be considered as the MLE estimator for the multivariate Student distribution with $\nu\rightarrow 0$~\cite[page 187]{Maronna2006}.
Due to the scale invariance property of $F(\bSigma)$ \begin{equation}F(\bSigma)=F(c\bSigma),\label{eq:scale_invariance}\end{equation} we enforce the condition $\tr(\bSigma)=1$ in \eqref{eq:problem} for the uniqueness of the minimizer.

Since the multivariate Student distribution
\[
\frac{\Gamma[(\nu+D)/2]}{\Gamma(\nu/2)\nu^{D/2}\pi^{D/2}\sqrt{\det\Sigma}[1+\frac{1}{\nu}\bx^T\Sigma^{-1}\bx]^{(\nu+D)/2}}
\]
is heavy-tailed, Tyler's M-estimator is robust to outliers. Indeed, Tyler~\cite{Tyler1987} showed that it is the ``most robust" estimator of the scatter matrix of an elliptical distribution in the sense of minimizing the maximum asymptotic variance. Therefore, we also expect it to perform well in robust subspace recovery. Now we are ready to present our main results:

\begin{thm}\label{thm:exact_recovery}
If there exists a $d$-dimensional subspace $L_*$ such that \begin{equation}\frac{|\sX\cap L_*|}{|\sX|}>\frac{d}{D},\label{eq:exact_recovery_condition}\end{equation} and the points in the sets $\{\bP_{L_*}\bx:\bx\in\sX\cap L_*\}\subset \reals^d$ and $\{\bP_{L_*^\perp}\bx:\bx\in\sX\setminus L_*\}\subset \reals^{D-d}$ lie in general position respectively (i.e., any $k$-dimensional linear subspace contains at most $k$ points), then the sequence $\bSigma^{(k)}$ generated by~\eqref{eq:IRLS} converges to some $\hat{\bSigma}$ such that $\im(\hat{\bSigma})= L_*$.
\end{thm}

Theorem~\ref{thm:exact_recovery} is our main result on robust subspace recovery: if the inliers lie exactly on the subspace $L_*$ and the percentage of inliers is larger than $\dim(L_*)/D$, then with some other weak assumptions we can recover $L_*$ by the range of $\lim_{k\rightarrow\infty}\bSigma^{(k)}$. The requirement of ``general position'' is weak: for example, it holds almost surely when we sample inliers from a distribution $\mu_1$ in $L_*$ and outliers from a distribution $\mu_0$ in $\reals^D$, where $\mu_0(L)=0$ for any subspace $L\subset\reals^D$, and $\mu_1(L')=0$ for any subspace $L'\subset L_*$.

One may wonder about the stability of Tyler's M-estimator, that is, what if the inliers do not lie on the subspace $L_*$ exactly? We will show that the span of the top $\dim(L_*)$ eigenvectors of Tyler's M-estimator is stable to noise and recovers $L_*$ approximately in Theorem~\ref{thm:stability1}.

We remark that generally, Tyler's M-estimator has the following property (see~\cite[Theorem 2]{KT88} and \cite[Proposition 1(a)]{tyler2005}):
\begin{thm}\label{thm:uniqueness}
If for all linear subspaces $L$ we have \begin{equation}\label{eq:uniqueness_assumption}\text{$\frac{ |\sX\cap L | }{N}<\frac{\dim(L)}{D}$,}\end{equation} then the solution of \eqref{eq:problem} exists and is unique, and the sequence $\bSigma^{(k)}$ generated by~\eqref{eq:IRLS} converges to the unique solution of \eqref{eq:problem}.
\end{thm}


The condition \eqref{eq:uniqueness_assumption} is almost the ``complement'' of the condition \eqref{eq:exact_recovery_condition} in Theorem~\ref{thm:exact_recovery}. Therefore, these two theorems together reveal a phase transition phenomenon at $\frac{|\sX\cap L_*|}{|\sX|}=\frac{d}{D}$: with more inliers, Tyler's M-estimator becomes singular and its range recover the underlying subspace; with fewer inliers, Tyler's M-estimator is full-rank and does not have the  property of exact subspace recovery. Additionally, Theorem 3 in \cite{KT88} also indicates that the existence of full-ranked Tyler's M-estimator implies   \eqref{eq:uniqueness_assumption}, which complements Theorem~\ref{thm:uniqueness} from a different direction.

\subsection{Previous works}
Robust subspace recovery has been studied in many works before. In particular, some works try to fit the linear model by PCA after removing possible outliers~\cite{Torre03aframework,Xu2010_highdimensional}. However, they lack strong theoretical guarantees: the method in~\cite{Torre03aframework} minimizes a nonconvex objective function by a heuristic iterative reweighted algorithm, which has no guarantee of the convergence to the minimizer. The theory in~\cite{Xu2010_highdimensional} only guarantees exact recovery of the subspace when the percentage of of outliers converges to $0$ asymptotically.
%

Some recent works on robust linear estimation~\cite{Xu2010,McCoy2011,Zhang2011,haystack2012} apply the tool of convex optimization and  provide conditions for exact subspace recovery (similar to Theorem~\ref{thm:exact_recovery}) as guarantee of performance. In particular, this work is related to the algorithm proposed in~\cite{Zhang2011}, which is given by the iterative procedure
\begin{equation}\label{eq:IRLS0}
\bQ^{(k+1)}=\big(\sum_{\bx\in\sX}\frac{\bx\bx^T}{\|\bQ^{(k)}\bx\|}\big)^{-1}/\trace\Big(\big(\sum_{\bx\in\sX}\frac{\bx\bx^T}{\|\bQ^{(k)}\bx\|}\big)^{-1}\Big).
\end{equation}
One can consider $\bQ$ in \eqref{eq:IRLS0} as ``inverse covariance'' and \eqref{eq:IRLS0} is equivalent to the procedure (up to a scaling)
\begin{equation}\label{eq:IRLS1}
\bSigma^{(k+1)}=\sum_{\bx\in\sX}\frac{\bx\bx^T}{\|\bSigma^{(k)\, -1}\bx\|}/\trace\Big(\sum_{\bx\in\sX}\frac{\bx\bx^T}{\|\bSigma^{(k)\, -1}\bx\|}\Big).
\end{equation}
Then it is clear that the difference between \eqref{eq:IRLS0} and \eqref{eq:IRLS} lies in the choice of the denominator of $\bx_i\bx_i^T$, i.e., the weight of each data point in the iterative procedure.

Compared to \eqref{eq:IRLS0}, the algorithm in \cite{haystack2012} have an additional step of thresholding the eigenvalues of $\bQ^{(k)}$ in each iteration of \eqref{eq:IRLS0}, which leads to a stronger theoretical guarantee on subspace recovery and a higher computational cost in each iteration, due to the SVD decomposition of $\bQ^{(k)}$.

Similar to Theorem~\ref{thm:exact_recovery}, these convex methods give theoretical guarantees on exact subspace recovery, and the conditions usually assume ``incoherence conditions'' that require the inliers to be spread out on  $L_*$~\cite[Theorem 1]{Xu2010}, ~\cite[(6)(7)]{Zhang2011}, or probabilistic distributions of inliers and outliers~\cite[Theorem 1.1]{haystack2012}. In comparison, our condition \eqref{eq:exact_recovery_condition} is much simpler and usually less restrictive. For example, \cite[Theorem 1.1]{haystack2012} shows exact recovery for the haystack model (i.e., inliers sampled from $N(\b0,\Proj_{L_*})$, outliers sampled from $N(\b0,\bI_D)$) with probability $1-4e^{\beta d}$ if
\[
\frac{|\sX\cap L_*|}{d}\geq C_1+C_2\beta +C_3 (\frac{|\sX\setminus L_*|}{D}+1+4\beta),
\]
where $C_1\approx 13, C_2\approx 7, C_3\approx  16$. Therefore, our condition \eqref{eq:exact_recovery_condition} is less restrictive due to these factors. Additionally, as shown later in Section~\ref{sec:face}, Tyler's M-estimator shows stronger robustness to outliers than the competitive methods empirically.


This superiority of Tyler's M-estimator has a theoretical guarantee from computational complexity theory: Hardt and Moitra~\cite{Hardt2012} studied the problem of robust subspace recovery and showed that it is small set expansion hard to recover a $d$-dimensional subspace with fewer than $(1-\eps)d/D\cdot N$ points, which is the threshold obtained by Tyler's M-estimator. It is conjectured that small set expansion might be NP-hard.


\subsection{Structure of this paper}
The paper is organized as follows: In Section~\ref{sec:background}, we introduce the background on the geometry of $\PSD(D)$ and the geodesic convexity. Then we prove Theorems~\ref{thm:uniqueness} and~\ref{thm:exact_recovery} and discuss the stability of subspace recovery by Tyler's M-estimator in Section~\ref{sec:main}.
Finally, we perform simulations to verify Theorem~\ref{thm:exact_recovery} and show the performance of Tyler's M-estimator on simulated and real data sets in Section~\ref{sec:experiments}. Technical proofs are shown in the Appendix.

\section{Preliminaries}\label{sec:background}
Our analysis of $F(\bSigma)$ relies on the property of geodesic convexity and the geometry of $\PSD(D)$. To make this paper self-contained, in Section~\ref{sec:background_spsd} we present a brief
summary of the geometry of $\PSD(D)$ and in Section~\ref{sec:background_convex} we introduce the definition of geodesic convexity. For more details on the geometry of $\PSD(D)$ and geodesic convexity, we refer the reader to \cite{bhatia2007positive,udriste1994convex}.
%

\subsection{Metric and geodesic on $\PSD(D)$}\label{sec:background_spsd}
The metric of $\PSD(D)$ has been studied in various fields. Interestingly, the trace metric in differential geometry~\cite[pg 326]{Lang99}, natural metric in symmetric cone~\cite{faraut1994analysis,Bonnabel09}, affine-invariant metric~\cite{Xavier06}, and the metric given by Fisher information matrix for Gaussian covariance matrix estimation~\cite{Smith05} give the same metric on $\PSD(D)$, which is defined by:
\begin{equation}
\dist(\bSigma_1,\bSigma_2)=\|\log(\bSigma_1^{-\frac{1}{2}}\bSigma_2\bSigma_1^{-\frac{1}{2}})\|_F,
\end{equation}
and the unique geodesic connecting $\bSigma_1$ and $\bSigma_2$ is given by~\cite[(6.11)]{bhatia2007positive}:
\begin{equation}\label{eq:geodesic}
\gamma_{\bSigma_1\bSigma_2}(t)=\bSigma_1^{\frac{1}{2}}(\bSigma_1^{-\frac{1}{2}}\bSigma_2\bSigma_1^{-\frac{1}{2}})^t\bSigma_1^{\frac{1}{2}}.
\end{equation}
It follows that the midpoint of $\bSigma_1$ and $\bSigma_2$ is $\gamma_{\bSigma_1\bSigma_2}(\frac{1}{2})=\bSigma_1^{\frac{1}{2}}(\bSigma_1^{-\frac{1}{2}}\bSigma_2\bSigma_1^{-\frac{1}{2}})^\frac{1}{2}\bSigma_1^\frac{1}{2}$.

\subsection{Geodesic convexity}\label{sec:background_convex}
Geodesic convexity is a generalization of the convexity from Euclidean space to Riemannian manifolds~\cite[Chapter 3.2]{udriste1994convex}. Given a Riemannian manifold $\mathcal{M}$ and a set $\mathcal{A}\subset \mathcal{M}$, a function $f: \mathcal{A}\rightarrow \reals$ is geodesically convex, if every geodesic $\gamma_{xy}$ of $\mathcal{M}$ with endpoints $x,y\in\mathcal{A}$ (i.e., $\gamma_{xy}$ is a function from $[0,1]$ to $\mathcal{M}$ with $\gamma_{xy}(0)=x$ and $\gamma_{xy}(1)=y$) lies in $\mathcal{A}$, and
  \begin{equation}\text{$f(\gamma_{xy}(t))\leq (1-t)f(x)+t f(y)$\,\, for any $x,y\in\mathcal{A}$ and $0< t< 1$}\label{eq:convex_defi}.\end{equation}



Following the proof of~\cite[Theorem 1.1.4]{niculescu2006convex}, for a continuous function, the geodesic midpoint convexity is equivalent to the geodesic convexity:
\begin{lemma}\label{lemma:convexity_equivalency}
Let $f:\mathcal{A}\rightarrow \reals$ be a continuous function. If
\begin{equation}\text{$f(\gamma_{xy}(\frac{1}{2})) \leq \frac{f(x)+ f(y)}{2}$ \,\,\,for any $x\neq y\in\mathcal{A}$}\label{eq:convex_defi3}\end{equation}
then $f$ is a geodesically convex function.
\end{lemma}

\section{The proof of main results}\label{sec:main}

In this section, we study the properties of the objective function $F(\bSigma)$ and the algorithm in~\eqref{eq:IRLS}, and prove Theorems~\ref{thm:uniqueness} and~\ref{thm:exact_recovery}. We first present the proof of Theorem~\ref{thm:uniqueness} since the proof of Theorems~\ref{thm:exact_recovery} is based on it. While parts of the proof of Theorem~\ref{thm:uniqueness} have appeared in previous works, we include them for the completeness of the paper. We also discuss an implementation issue in Section~\ref{sec:implementation} and the stability of subspace recovery in Section~\ref{sec:stability}.


The proof of Theorem~\ref{thm:uniqueness} depends on the following two lemmas. In particular, Lemma~\ref{lemma:property_objective_function} guarantees the uniqueness of the solution and Lemma~\ref{lemma:singular_infinity} guarantees the existence of the solution. While \eqref{eq:convexity} has been proved in~\cite[Proposition 1]{Wiesel2012}, we additionally prove the important property of strict convexity, which gives the uniqueness of the solution to \eqref{eq:problem}. The proof of Lemma~\ref{lemma:property_objective_function} is deferred to Section~\ref{sec:property}. Lemma~\ref{lemma:singular_infinity} is a restatement of \cite[Theorem 1]{KT88}, and we refer the reader to it for the proof.

\begin{lemma}\label{lemma:property_objective_function}
$F(\bSigma)$ is geodesically convex on the manifold $\PSD(D)$. That is, for any $\bSigma_1$ and $\bSigma_2\in\PSD(D)$, we have
\begin{equation}\label{eq:convexity}
F(\bSigma_1)+F(\bSigma_2)\geq 2 F(\bSigma_1^\frac{1}{2}(\bSigma_1^{-\frac{1}{2}}\bSigma_2\bSigma_1^{-\frac{1}{2}})^\frac{1}{2}\bSigma_1^\frac{1}{2}).
\end{equation}

When $\mathrm{span}\{\sX\}=\mathbb{R}^D$, the equality in \eqref{eq:convexity} holds if and only if $\bSigma_1=c\bSigma_2$.
\end{lemma}

\begin{lemma}\label{lemma:singular_infinity}
Under the condition~\eqref{eq:uniqueness_assumption}, we have
\begin{equation}\text{$F(\bSigma)\rightarrow\infty$ as $\lambda_{\min}(\bSigma)\rightarrow 0$.}\label{eq:convergence_singular}\end{equation}
Here $\lambda_{\min}(\bSigma)$ is the smallest eigenvalue of $\bSigma$.
\end{lemma}

With Lemmas~\ref{lemma:property_objective_function} and \ref{lemma:singular_infinity} we are ready to prove the uniqueness and existence of the solution to \eqref{eq:problem}.

\begin{proof}
We first prove the uniqueness of the solution to \eqref{eq:problem}. If $\bSigma_1\neq\bSigma_2$ are both solutions to \eqref{eq:problem}, then applying \eqref{eq:convexity} and the scale invariance in \eqref{eq:scale_invariance}, we have \[F(\bSigma_3)\leq F(\bSigma_1)=F(\bSigma_2),\,\,\,\text{for}\]
\[
\bSigma_3=\frac{\bSigma_1^\frac{1}{2}(\bSigma_1^{-\frac{1}{2}}\bSigma_2\bSigma_1^{-\frac{1}{2}})^\frac{1}{2}\bSigma_1^\frac{1}{2}}{\tr\Big(\bSigma_1^\frac{1}{2}(\bSigma_1^{-\frac{1}{2}}\bSigma_2\bSigma_1^{-\frac{1}{2}})^\frac{1}{2}\bSigma_1^\frac{1}{2}\Big)}.
\]
Since $\bSigma_1$ and $\bSigma_2$ are both minimizers to $F(\bSigma)$, we have $F(\bSigma_3)=F(\bSigma_1)=F(\bSigma_1)$. Applying the condition of achieving equality in \eqref{eq:convexity} (the assumption $\mathrm{span}\{\sX\}=\mathbb{R}^D$ in Lemma~\ref{lemma:property_objective_function} holds; otherwise \eqref{eq:uniqueness_assumption} does not hold for $L=\mathrm{span}\{\sX\}$), we have $\bSigma_1=c\bSigma_2$. Since $\tr(\bSigma_1)=\tr(\bSigma_2)=1$, we have $\bSigma_1=\bSigma_2$, which is a contradiction to the previous assumption. Therefore, we proved the uniqueness of the solution to \eqref{eq:problem}.

Now we prove the existence of the solution. First, there exists a sequence $\{\bSigma_i\}_{i\geq 1}\subset\{\bSigma\in\PSD(D):\tr(\bSigma)=1\}$ such that $F(\bSigma_i)$ converges to $\inf_{\tr(\bSigma)=1,\bSigma\in\PSD(D)}F(\bSigma)$. By the compactness of the set $\{\bSigma\in\SPSD(D):\tr(\bSigma)=1\}$, there is a converging subsequence of $\{\bSigma_i\}$, and by Lemma~\ref{lemma:singular_infinity} this subsequence does not converge to a singular matrix and therefore, the subsequence converges to some matrix $\bSigma_0\in\PSD(D)$. By the continuity of $F(\bSigma)$ we have $F(\bSigma_0)=\inf_{\tr(\bSigma)=1,\bSigma\in\PSD(D)}F(\bSigma)$ and therefore $\bSigma_0$ is a solution to \eqref{eq:problem}.
\end{proof}


\subsection{Theorem~\ref{thm:uniqueness}: Convergence of the algorithm}

In this section, we prove the convergence of the sequence $\bSigma^{(k)}$ generated by~\eqref{eq:IRLS} under the assumption~\eqref{eq:uniqueness_assumption}. Similar to~\cite[Section II]{Wiesel2012}, it uses the majorization-minimization argument~\cite{MM_tutorial2004}. However our analysis is more complete since it proves the convergence of the sequence $\bSigma^{(k)}$, while the argument in \cite{Wiesel2012} only gives the convergence of the objective function $F(\bSigma^{(k)})$.

\begin{proof}
For simplicity we define the operator
$T:\SPSD(D)\rightarrow\SPSD(D)$ as
\begin{equation}\label{eq:def_T}T(\bSigma)
=\sum_{\bx\in\sX}\frac{\bx\bx^T}{\bx^T\bSigma^{\,-1}\bx}/
\trace(\sum_{\bx\in\sX}\frac{\bx\bx^T}{\bx^T\bSigma^{\,-1}\bx}).\end{equation}

First we will prove that the operator $T$ is monotone with respect to the objective function $F$: $F(T(\bSigma))\leq F(\bSigma)$ and the equality holds for $\bSigma\in\PSD(D)$ if and only if $T(\bSigma)=\bSigma$.

We prove it by constructing the following majorization function over $F(\bSigma)$:
\begin{equation}
G(\bSigma,\bSigma^*)=\left\langle
\frac{1}{N}\sum_{\bx\in\sX}\frac{\bx^T\bx}{\bx^T\bSigma^{*\,-1}\bx},\bSigma^{-1}
\right\rangle+\frac{1}{D}\log\det(\bSigma)+C,
\end{equation}
where $C$ is chosen such that $G(\bSigma^*,\bSigma^*)=F(\bSigma^*)$. The fact  \[G(\bSigma,\bSigma^*)\geq F(\bSigma)\] can be proved by checking the first and the second derivatives of $G(\bSigma,\bSigma^*)-F(\bSigma)$ with respect to $\bSigma^{-1}$.

It is easy to verify the unique minimizer of $G(\bSigma,\bSigma^*)$ is
\[
\tilde{\bSigma}=\frac{D}{N}\sum_{\bx\in\sX}\frac{\bx^T\bx}{\bx^T\bSigma^{*\,-1}\bx},
\]
which is a scaled version of $T(\bSigma^*)$. Then we prove the monotonicity of $T$ as follows:
\begin{equation}\label{eq:monotone}
F(T(\bSigma^*))=F(\tilde{\bSigma})\leq G(\tilde{\bSigma},\bSigma^*)
\leq G(\bSigma^*,\bSigma^*)=F(\bSigma^*).
\end{equation}
Because of the uniqueness of the minimizer of $G(\bSigma,\bSigma^*)$, the equality in the second inequality of \eqref{eq:monotone} holds only when $\tilde{\bSigma}=\bSigma^*$. Since $\tilde{\bSigma}=c T(\bSigma^*)$ and $\tr(\bSigma^*)=\tr(T(\bSigma^*))=1$,
the equality in \eqref{eq:monotone} holds if and only if $T(\bSigma^*)=\bSigma^*$.

Therefore the sequence $F(\bSigma^{(k)})$ is monotone, and any accumulation point of the sequence $\{\bSigma^{(k)}\}$ (denoted by $\hat{\bSigma}$) satisfies $F(T(\hat{\bSigma}))=F(\hat{\bSigma})$. Applying the condition of achieving equality in \eqref{eq:monotone}, we have $T(\hat{\bSigma})=\hat{\bSigma}$, which is equivalent to
\begin{equation}\label{eq:derivative}
\hat{\bSigma}\sum_{\bx\in\sX}\frac{\bx^T\bx}{\bx^T\hat{\bSigma}^{-1}\bx}=c\bI,\,\,\,\text{for some $c\in\reals$}.
\end{equation}
Let $\bA=\log(\bSigma^{-1})$, applying $\log\det(\bSigma)=-\tr(\bA)$ and $\frac{\di}{\di \bA}\exp(\bA)=\exp(\bA)$, the derivative of $F(\bSigma)$ with respect to $\bA$ is
\[
\frac{\di}{\di\bA}F(\bSigma)=\frac{1}{N}\bSigma^{-1}\,\sum_{\bx\in\sX}\frac{\bx^T\bx}{\bx^T\hat{\bSigma}^{-1}\bx}
-\frac{1}{D}\bI.
\]
Since $\{\bA:\bA=\log(\bSigma^{-1}),\,\,\text{where $\det(\bSigma)=1$}\}=\{\bA:\tr(\bA)=1\}$, applying \eqref{eq:derivative}, all directional derivatives of $\frac{\di}{\di\bA}F(\bSigma)\big|_{\bSigma=c_0\hat{\bSigma}}$ in the set $\{\bSigma:\det(\bSigma)=1\}$ are $0$, where $c_0$ is a number chosen such that $\det(c_0\hat{\bSigma})=1$. Since both the set $\{\bSigma:\det(\bSigma)=1\}$ and $F(\bSigma)$ are geodesically convex, $c_0\hat{\bSigma}$ is the unique minimizer of $F(\bSigma)$ in the set $\{\bSigma:\det(\bSigma)=1\}$. Applying the scale invariance
 of $F(\bSigma)$ in~\eqref{eq:scale_invariance}, $\hat{\bSigma}$ is the unique solution in the set $\{\bSigma:\tr(\bSigma)=1\}$, i.e., it is the unique solution to \eqref{eq:problem}.
\end{proof}
\subsection{Proof of Theorem~\ref{thm:exact_recovery}}
First of all the algorithm can be written as 
\[
\Sigma^{(k)}=\frac{\sum_{i=1}^Nw_i^{(k)}\bx_i\bx_i^T}{\tr\left(\sum_{i=1}^Nw_i^{(k)}\bx_i\bx_i^T\right)},
\]
where the update formula of $w_i^{(k)}$ given by 
\[
w_i^{(k+1)}=\frac{1}{\bx_i^T(\sum_{i=1}^Nw_i^{(k)}\bx_i\bx_i^T)^{-1}\bx_i}.
\]
We denote the set of outliers by $\sX_0=\sX\setminus L_*$ and set of inliers by $\sX_1=\sX\cap L_*$. We let $\mathcal{I}_1=\{1\leq i\leq N: \bx_i\in L_*\}$ be the set of indices of inliers and $\mathcal{I}_0=\{1\leq i\leq N: \bx_i\not\in L_*\}$ be the the set of the indices of outliers, $N_1=|\sX_1|$, $N_0=|\sX_0|$. We denote an elementwise linear transformation $\bA$ on the set $\sX$ by $\bA(\sX)=\{\bA\bx:\bx\in\sX\}$ and the solutions of \eqref{eq:problem} for the set $\sX$ by $\bSigma_*(\sX)$. WLOG we may assume that $\bSigma_*(\bP_{L_*}(\sX_1))=\bI_d/d$ and $\bSigma_*(\bP_{L_*^\perp}(\sX_0))=\bI_{D-d}/(D-d)$. Since TME is invariant to scaling of $\bx_i$, we may assume that $i\in\mathcal{I}_1$, $\|\bx_i\|=1$. Then  $\bSigma_*(\bP_{L_*}(\sX_1))=\bI_d/d$ implies
\[
\sum_{i\in\mathcal{I}_1}{\bx_i\bx_i^T}=\frac{N_1}{d}\Proj_{L_*}.
\]
Similarly, for all $i\in\mathcal{I}_0$, we assume $\|P_{L_*^\perp}\bx_i\|=1$ and $\bSigma_*(\bP_{L_*^\perp}(\sX_0))=\bI_{D-d}/(D-d)$ implies
\[
\sum_{i\in\mathcal{I}_0}{P_{L_*^\perp}\bx_i\bx_i^TP_{L_*^\perp}^T}=\frac{N_0}{D-d}\Proj_{L_*^\perp}.
\]

With these assumptions, for any $i\in\mathcal{I}_1$, 
\begin{align*}
&w_i^{(k+1)}\geq \frac{1}{\bx_i^T(\sum_{i \in \mathcal{I}_1}w_i^{(k)}\bx_i\bx_i^T)^{-1}\bx_i}\geq \frac{\min_{i \in \mathcal{I}_1} w_i^{(k)}}{\bx_i^T(\sum_{i \in \mathcal{I}_1}\bx_i\bx_i^T)^{-1}\bx_i}= \frac{\min_{i \in \mathcal{I}_1} w_i^{(k)}}{\bx_i^T(\frac{N_1}{d}\Proj_{L_*})^{-1}\bx_i}\\
= & \frac{N_1}{d}\min_{i \in \mathcal{I}_1}w_i^{(k)}
\end{align*}
On the other hand, for any $i\in\mathcal{I}_0$,
\begin{align*}
&w_i^{(k+1)}\leq \frac{1}{\bx_i^T(\infty\cdot \Proj_{L_*}+\sum_{i \in \mathcal{I}_0}w_i^{(k)}\bx_i\bx_i^T)^{-1}\bx_i}=\frac{1}{\bx_i^TP_{L_*^\perp}^T(\sum_{i \in \mathcal{I}_0}w_i^{(k)}P_{L_*^\perp}\bx_i\bx_i^TP_{L_*^\perp}^T)^{-1}P_{L_*^\perp}\bx_i}\\\leq& \frac{\max_{i \in \mathcal{I}_0}w_i^{(k)}}{\bx_i^TP_{L_*^\perp}^T(\sum_{i \in \mathcal{I}_0}P_{L_*^\perp}\bx_i\bx_i^TP_{L_*^\perp}^T)^{-1}P_{L_*^\perp}\bx_i}= \frac{N_0}{D-d}\max_{i \in \mathcal{I}_0}w_i^{(k)}
\end{align*}

As a result, we have
\[
\frac{\min_{i \in \mathcal{I}_1}w_i^{(k+1)}}{\max_{i \in \mathcal{I}_0}w_i^{(k+1)}}\geq \frac{N_1(D-d)}{N_0d}\frac{\min_{i \in \mathcal{I}_1}w_i^{(k)}}{\max_{i \in \mathcal{I}_0}w_i^{(k)}}\geq \cdots\geq \left(\frac{N_1(D-d)}{N_0d}\right)^k\frac{\min_{i \in \mathcal{I}_1}w_i^{(1)}}{\max_{i \in \mathcal{I}_0}w_i^{(1)}}.
\]
Since $\frac{N_1(D-d)}{N_0d}>1$, as $k\rightarrow\infty$, compared to the weights of the inliers, the weights of the outliers decrease exponentially  and $\Sigma^{(k)}$ becomes singular.

As a result, the TME algorithm of $\sX$ becomes the TME algorithm on $\sX_1$ and converges to $\Proj_{L_*}/d$.

%
%
%

\subsection{Implementation issues}\label{sec:implementation}
%
%
%
Careful readers may notice that the algorithm~\eqref{eq:IRLS} breaks down if $\Sigma^{(k)}$ is singular and wonder if this could be problematic in implementation. Here we make several remarks on this issue.

 First, for general data sets, the condition \eqref{eq:uniqueness_assumption} almost always holds, and the algorithm~\eqref{eq:IRLS} does not break down. Applying~\cite[Theorem 1]{KT88}, $L(\bSigma)\rightarrow\infty$ as $\lambda_{\min}(\bSigma)$ approaches zero. Since $L(\Sigma^{(k)})$ is non-increasing (as shown in the proof of Theorem~\ref{thm:uniqueness}), $\lambda_{\min}(\Sigma^{(k)})$ is bounded from below. Therefore, the inversion of  $\Sigma^{(k)}$ in \eqref{eq:IRLS} has no numerical issue.


Second, even if the condition \eqref{eq:uniqueness_assumption} does not hold and $\Sigma^{(k)}$ becomes numerically singular for some large $k$ (that is, $\Sigma^{(k)}$ has very small eigenvalues), we claim that $\Sigma^{(k)}$ can be used to recover $L_*$. Based on this claim, in our implementation, we stop the algorithm when the algorithm shows instability, that is, when $\bSigma^{(k)}$ is numerically singular. Then we recover the underlying subspace by the span of the top eigenvectors of $\bSigma^{(k)}$.

The argument for the claim is as follows. Assume that $(\bSigma)^{-1}$ is numerically unstable when $\lambda_{\min}(\bSigma)<\eps$, then since
\[
\frac{\lambda_{\max}(T(\bSigma))}
{\lambda_{\min}(T(\bSigma))}
=\frac{\lambda_{\max}(\sum_{\bx\in\sX}\frac{\bx\bx^T}{\bx^T\bSigma^{\,-1}\bx} )}
{\lambda_{\min}(\sum_{\bx\in\sX}\frac{\bx\bx^T}{\bx^T\bSigma^{\,-1}\bx} )}
\leq
\frac{\lambda_{\max}(\bSigma)\lambda_{\max}(\sum_{\bx\in\sX}\frac{\bx\bx^T}{\|\bx\|^2} )}
{\lambda_{\min}(\bSigma)\lambda_{\min}(\sum_{\bx\in\sX}\frac{\bx\bx^T}{\|\bx\|^2} )},
\]
we have
\[
\frac{\lambda_{\max}(\bSigma^{(k)})}
{\lambda_{\min}(\bSigma^{(k)})}
\leq C_2\,C_3^{k-1},\,\,\text{where $C_2=\frac{\lambda_{\max}(\bSigma^{(1)})}
{\lambda_{\min}(\bSigma^{(1)})}$, $C_3=\Big(\frac{\lambda_{\max}(\sum_{\bx\in\sX}\frac{\bx\bx^T}{\|\bx\|^2} )}
{\lambda_{\min}(\sum_{\bx\in\sX}\frac{\bx\bx^T}{\|\bx\|^2} )}\Big)$}.
\]
Therefore the algorithm is stable for the first $k_1$ iterations, where $k_1=\log(D\,C_2/\eps)/\log(C_3)$. And by \eqref{eq:ratio2} we know that
\[
\frac{\lambda_{\min}(\bP_{L_*}^T\bSigma^{(k_1)}\bP_{L_*})}{\lambda_{\max}(\bP_{L_*^\perp}^T \bSigma^{(k_1)}\bP_{L_*^\perp})}\geq \alpha^{k_1-1}\,\frac{\lambda_{\min}(\bP_{L_*}^T\bSigma^{(1)}\bP_{L_*})}{\lambda_{\max}(\bP_{L_*^\perp}^T \bSigma^{(1)}\bP_{L_*^\perp})},
\]
so when the precision $\eps$ is sufficiently small,
\begin{equation}\label{eq:eigengap1}\text{$\frac{\lambda_{\min}(\bP_{L_*}^T\bSigma^{(k_1)}\bP_{L_*})}{\lambda_{\max}(\bP_{L_*^\perp}^T \bSigma^{(k_1)}\bP_{L_*^\perp})}$ is sufficiently large, and $\frac{\lambda_{\max}(\bP_{L_*}^T \bSigma^{(k_1)}\bP_{L_*})}{\lambda_{\min}(\bP_{L_*}^T\bSigma^{(k_1)}\bP_{L_*})}$ is bounded above due to \eqref{eq:max_min_rate3}.}\end{equation} By the Courant-Fischer min-max theorem~\cite[Theorem 1.3.2]{tao_ran_mat_book}, the $d$-th eigenvalue of $\bSigma^{(k_1)}$  is larger than $\lambda_{\max}(\bP_{L_*^\perp}^T \bSigma^{(k_1)}\bP_{L_*^\perp})$ and the $(d+1)$-th eigenvalues of $\bSigma^{(k_1)}$ is smaller than $\lambda_{\min}(\bP_{L_*}^T\bSigma^{(k_1)}\bP_{L_*})$. And by \cite[Lemma 3.2]{Carlen2010},
\[
\bSigma^{(k_1)}=\left[ \begin{array}{cc}
(\bP_{L_*}^T \bSigma^{(k_1)}\bP_{L_*})^{\frac{1}{2}}  & 0 \\
0 & (\bP_{L_*^\perp}^T \bSigma^{(k_1)}\bP_{L_*^\perp} )^{\frac{1}{2}} \end{array} \right]\left[ \begin{array}{cc}
\bI & \bU^T \\
\bU & \bI \end{array} \right]\left[ \begin{array}{cc}(\bP_{L_*}^T \bSigma^{(k_1)}\bP_{L_*})^{\frac{1}{2}}  & 0 \\
0 & (\bP_{L_*^\perp}^T \bSigma^{(k_1)}\bP_{L_*^\perp})^{\frac{1}{2}} \end{array} \right],
\]
where $\bI-\bU^T\bU$ and $\bI-\bU\bU^T$ are positive definite. Therefore $\lambda_{\max}(\bSigma^{(k_1)})\leq 2\lambda_{\max}((\bP_{L_*}^T \bSigma^{(k_1)}\bP_{L_*}))$. Combining it with \eqref{eq:eigengap1}, we obtain that  $\bSigma^{(k_1)}$ has a clear eigengap between the $d$-th eigenvalue and $(d+1)$-th eigenvalue, where $d=\dim(L_*)$. By the Davis-Kahan theorem, the span of the top $d$ eigenvectors of $\bSigma^{(k_1)}$ is a good approximation of $L_*$.

\subsection{Stability of subspace recovery}\label{sec:stability}
In this section, we analyze the stability of subspace recovery by Tyler's M-estimator, when data set consists of a clean component and a component of noise, and the clean component satisfies the assumptions in  Theorem~\ref{thm:exact_recovery}.


We first show that $\hat{\bSigma}=\lim_{k\rightarrow\infty}\bSigma^{(k)}$ is not robust to noise with the following example. Assuming that $\sX=\{\bx_1,\bx_2,\cdots,\bx_{10}\}\subset\reals^3$ and $\bx_1, \bx_2, \cdots, \bx_8\in L_*$ for a two-dimensional subspace $L_*$. Since there are $8$ inliers and $8/10>2/3$, by the proof of Theorem~\ref{thm:exact_recovery}, the algorithm converges to $\bP_{L^*}\Sigma_*(\{\bP_{L_*}\bx_1,\bP_{L_*}\bx_2,\cdots,\bP_{L_*}\bx_8\})\bP_{L^*}^T$.

Now we add an arbitrarily small noise to $\bx_8$ and keep other points unchanged. Now there are $7$ inliers and $7/10>2/3$, so following the same argument, the algorithm converges to the different matrix $\bP_{L^*}\Sigma_*(\{\bP_{L_*}\bx_1,\bP_{L_*}\bx_2,\cdots,\bP_{L_*}\bx_7\})\bP_{L^*}^T$. That is, Tyler's M-estimator could be unstable to an arbitrary small noise.

While Tyler's M-estimator itself is unstable to small noise, we still have the following statement, which shows that Tyler's M-estimator is robust for the purpose of recovering subspace. Its proof is rather technical and is deferred to Section~\ref{sec:thm_stability1}.



%
%
%
%
%
\begin{thm}\label{thm:stability1}
Assume a data set $\sX=\{\bx_i\}_{i=1}^N$, $N>2D$, all points lie on the unit sphere, i.e., $\|\bx_i\|=1$ for all $1\leq i\leq N$, and $\{\bx_i\}_{i=1}^{N_1}$ lie approximately on a $d$-dimensional subspace $L_*$ in the sense that $\dist(\bx_i, L_*)<\eps$ for all $1\leq i\leq N_1$, and additionally we have
\begin{enumerate}\item  The percentage of the inliers is larger than $d/D$: $\alpha= \frac{N_1 D}{N d}>1$.
\item Data points do not concentrate around any subspace other than $L_*$: There exists constants $c_1, C_1$ and $\eta_0$ such that if a subspace $L$ approximately contains more than $(\dim(L) /D - c_1) N$ points in the sense that $|\{1\leq i\leq N:\dist(\bx_i,L)<\eta\}| > ( \dim(L) /D - c_1) N$, where $\eta<\eta_0$, then $\dim(L)\geq d$ and $L$ approximately contains $L_*$: $\|\bP_{L_*}-\bP_{P_{L}L_*}\|<C_1 \eta$,  where $P_{L}L_*$ is the subspace obtained by projecting all points in $L_*$ to $L$: $P_{L}L_* = \{\Proj_{L}\bx:\bx\in L_*\}$.
\item The set of outliers $\{\bx_i\}_{i=N_1+1}^N$ does not concentrate around any subspace in $\reals^D$: For any $m$-dimensional subspace $L$ containing $L_*$, $|\{N_1+1\leq i\leq N:\dist(\bx_i, L)<\eta\}|< \max(C_2\eta N, m-d )$.
\end{enumerate}.

Then there exists a constant $C_0>0$ such that when $\eps< \min( c_1\sqrt{C_0},\eta_0\sqrt{C_0}, 1/2)$, we have $\|\bP_{L_d}-\bP_{L_*}\|< \frac{\eps C_1}{\sqrt{C_0}}$. The parameter $C_0$ is specified later in \eqref{eq:beta} and \eqref{eq:C0} and does not depend on $N$.
\end{thm}

We may assume WLOG that $\|\bx_i\|=1$ since Tyler's M-estimator is invariant to the scaling of each data point, and the condition $N>2D$ is required such that the RHS of \eqref{eq:kappam} is positive. We note that the choice of the factor of $2$ is arbitrary and Theorem~\ref{thm:stability1} still holds (with a different choice of $C_0$) if we replace the condition by $N> c D$ for any other $c>1$ such as $c=1.1$.

Now we explain the three assumptions in the statement of Theorem~\ref{thm:stability1}. The first assumption is equivalent to \eqref{eq:exact_recovery_condition} and is necessary since this is a generalization of  Theorem~\ref{thm:exact_recovery} to the noisy case. Both the second and the third assumptions force the distribution of data points to be approximately uniform with the exception of the concentration around $L_*$. To see this point, let use consider the following model: the inliers $\{\bx_i\}_{i=1}^{N_1}$ are sampled uniformly from the unit sphere in $L_*$, and the outliers $\{\bx_i\}_{i=N_1+1}^N$ are sampled uniformly from the unit sphere in the ambient space $\mathbb{R}^D$. We state with proof that the assumptions are satisfied asymptotically with $c_1=\frac{d}{2D}$, $C_1=\frac{2D\,B(\frac{D-1}{2},\frac{1}{2})}{ \pi d }$, $\eta_0=\frac{\pi d }{2D\,B(\frac{D-1}{2},\frac{1}{2})}$ and $C_2=\frac{\pi}{B(\frac{D-1}{2},\frac{1}{2})}$, where $B$ represents the beta function.

The proof is divided in three steps. For the first step, we prove that the conditional number of $\hat{\bSigma}$ is large. Second, we will show that $\hat{\bSigma}$ has $d$ large eigenvalues and $D-d$ smaller eigenvalues. At last, we will show that the span of the $d$ large eigenvectors approximately recovers the subspace $L^*$.

\section{Numerical Experiments}\label{sec:experiments}
In this section, we run some simulations to investigate the empirical performance of this algorithm. We also show that Tyler's M-estimator outperforms other convex algorithms of robust PCA on a real data set.

\subsection{Model for simulation}\label{sec:model_simulation}
In Sections~\ref{sec:simulation_exact_recovery}-\ref{sec:simulation_noise},
 we apply the algorithm \eqref{eq:IRLS} to data sets generated from the following model.
  We choose a $d$-dimensional subspace $L_*$, sample $N_1$ points
  i.i.d. from the Gaussian distribution $N(0,\Proj_{L_*})$ on
   $L_*$, and sample $N_0$ outliers i.i.d. from the uniform distribution in the cube $[0,1]^D$. We use this distribution of outliers so that the outliers are anisotropic.
   In some experiments we also add a Gaussian noise $N(0,\eps^2\bI)$
   to each of the point.

\subsection{Exact recovery of the subspace}\label{sec:simulation_exact_recovery}
\begin{figure}[htbp]
\begin{center}
\includegraphics[width=.47\textwidth,height=.4\textwidth]{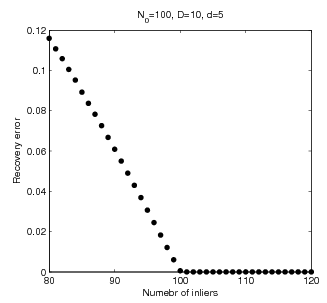}
\includegraphics[width=.47\textwidth,height=.4\textwidth]{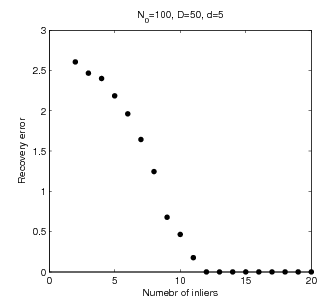}
\caption{\it The dependence on the number of inliers and recovery error: $x$-axis is the number of inlier and $y$-axis is the corresponding recovery. error\label{fig:exact_recovery}}
\end{center}
\end{figure}
In this section, we choose $D=10$ or $50$, $d=5$, $N_0=100$
 and different values of $N_1$ ($2$ to $20$ for $D=50$
 and $80$ to $120$ for $D=10$).
 The mean recovery error $\|\Proj_{\hat{L}}-\Proj_{L_*}\|_F$ over 20 experiments
is recorded in
 Figure~\ref{fig:exact_recovery}, where $\hat{L}$ is obtained by the span of top $d$ eigenvectors of Tyler's M-estimator and $L_*$ is the true underlying subspace.
 Theorem~\ref{thm:exact_recovery} guarantees exact subspace recovery, i.e., $\|\Proj_{\hat{L}}-\Proj_{L_*}\|_F=0$ for $N_1>100$ when $D=10$ and $N_1>10$ when $D=50$, and it is verified by this experiment.
 When $D=50$ and $N_1=11$ there is a small nonzero recovery error, which seems to contradict Theorem~\ref{thm:exact_recovery}, but we remark that when $D=50$ and $N_1=11$ the convergence is slow, and we stop the algorithm at the 1000-th iteration without the eventual convergence to the solution to \eqref{eq:problem}. We expect that the exact recovery of $L_*$ might require a large number of iterations.

\subsection{Convergence rate}\label{sec:simulation_convergence}
\begin{figure}
\begin{center}
\includegraphics[width=.47\textwidth,height=.4\textwidth]{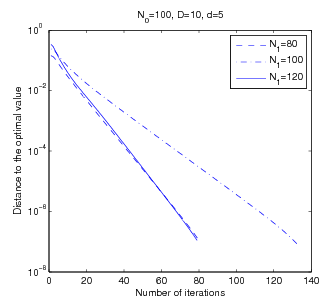}
\includegraphics[width=.47\textwidth,height=.4\textwidth]{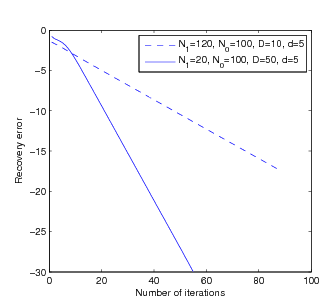}
\caption{\it Convergence rate for simulated data sets. See the text in Section~\ref{sec:simulation_convergence} for more details of the experiment.\label{fig:convergence_rate1}}
\end{center}
\end{figure}

In this section, we show that empirically the algorithm converges linearly. In the left figure in Figure~\ref{fig:convergence_rate1}, we show the convergence rate for simulated data sets with $D=10$, $d=5$, $N_0=100$ and $N_1=80,100,120$, and we add a Gaussian noise with $\eps=0.01$. The $x$-axis represents the number of iterations $k$ and the $y$-axis represents $\|\bSigma^{(k)}-\bSigma_*\|_F$. From the left figure in Figure~\ref{fig:convergence_rate1} we see that $\|\bSigma^{(k)}-\bSigma_*\|_F$ converges linearly. We also show a different convergent rate: we plot the error of recovered subspace if we use $L_k$, the span of first $d$ eigenvectors of $\bSigma^{(k)}$ to recover the underlying subspace. In particular, we plot $\bSigma^{(k)}$ $\|\Proj_{L_k}-\Proj_{L_*}\|_F$ with respect to the number of iterations $k$. We use the settings $(N_1,N_0,D,d)=(120,100,10,5)$ and $(20,100,50,5)$ and we do not add noise, so Theorem~\ref{thm:exact_recovery} predicts that $\|\Proj_{L_k}-\Proj_{L_*}\|_F$ converges to $0$.  From the right figure in Figure~\ref{fig:convergence_rate1} we see that the recovery error converges to $0$ and the rate of convergence is also linear.

%
\subsection{Robustness to noise}\label{sec:simulation_noise}
\begin{figure}
\begin{center}
\includegraphics[width=.5\textwidth,height=.3\textwidth]{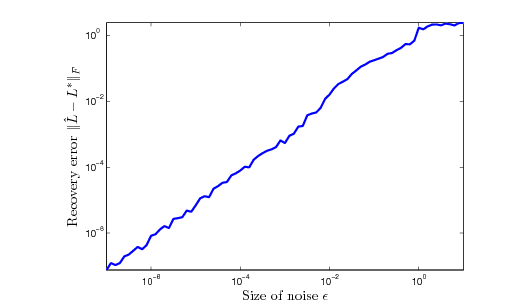}
\label{fig:noise}
\caption{\it Robustness to noise: the $x$-axis represents the size of Gaussian noise $\eps$, and the $y$-axis represents the recovery error. }
\end{center}
\end{figure}

In this section we investigate the robustness of Tyler's M-estimator to noise by simulated data sets with $(N_1,N_0,D,d)=(120,100,10,5)$ and various noise sizes $\eps$. We use this setting since when $\eps=0$, the subspace is recovered exactly and the recovery error is $0$. We record the recovery error in Figure~\ref{fig:noise} with respect to the size of noise $\eps$. In this experiment, the recovery error depends linearly on the size of noise, which is same rate as in the statement of Theorem~\ref{thm:stability1}. 

\subsection{Faces in a Crowd}\label{sec:face}
In this section we test Tyler's M-estimator on the experiment of ``Faces in a Crowd'' described in~\cite[Section 5.4]{haystack2012}.

The purpose of this experiment is to show that our algorithm recovers the structure of face images robustly. Linear modeling is applicable here since the images of the faces of the same person lies around a nine-dimensional subspace~\cite{Basri03}. In this experiment we learn the subspace from a data set that contains 32 face images of a person from the Extended Yale Face
Database~\cite{KCLee05} and 400 random images from the BACKGROUND/Google folder of
the Caltech101 database~\cite{caltech101}. The images are converted to grayscale and downsampled to $20\times 20$. We preprocess the images by subtracting their Euclidean median, and use the span of top eigenvectors of the solution to \eqref{eq:IRLS} to obtain a $9$-dimensional subspace, and then we use $32$ other images from the same person to test the ``goodness'' of the recovered subspaces, and we expect clearer images from the better methods.

\begin{figure}
\begin{center}
\includegraphics[width=.5\textwidth,height=.4\textwidth]{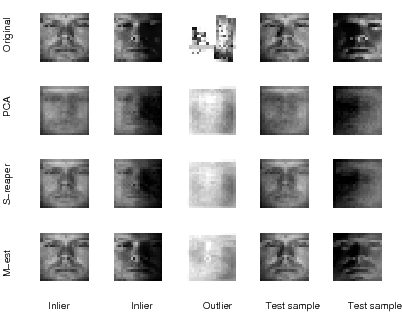}
\label{fig:face1}
\caption{\it The projection of images to the fitted subspace. }
\end{center}
\end{figure}

 This experiment is also used in~\cite[Section 5.4]{haystack2012}, therefore we only compare Tyler's M-estimator with S-Reaper, which has been shown to outperform spherical PCA, LLD and Reaper algorithms. PCA algorithm is still included for comparison since it is the basic method of linear modeling. Figure~\ref{fig:face1} shows five images and their projections to the 9-dimensional subspace fitted by PCA, S-reaper and Tyler's M-estimator (which is labeled as ``M-estimator'') respectively, and it shows that Tyler's M-estimator visually performs better than S-Reaper, especially for the test images. This observation can also be quantitatively verified by checking the distances of 32 test images to the fitted subspace by PCA, S-reaper and Tyler's M-estimator. The subspace generated by Tyler's M-estimator has smaller distances to the test images, which explain the better performance of Tyler's M-estimator in Figure~\ref{fig:face1}.

 Besides, in this experiment Tyler's M-estimator performs much faster than S-Reaper; Tyler's M-estimator costs 4.4 seconds on a machine with Intel Core 2 Duo CPU at 3.00GHz and 6GB memory, while S-reaper cost 40 seconds. The difference of the running time might be due to the additional eigenvalue decomposition step in each iteration of the S-Reaper algorithm.
\begin{figure}
\begin{center}
\includegraphics[width=.5\textwidth,height=.3\textwidth]{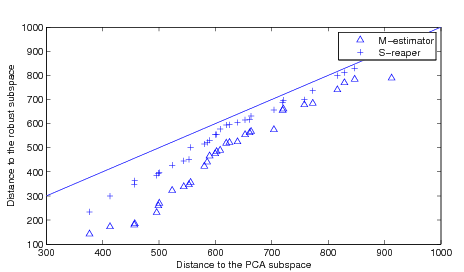}
\label{fig:face2}
\caption{\it Ordered distances of the 32 test images to the fitted $9$-dimensional subspaces by Tyler's M-estimator, S-reaper and PCA. }
\end{center}
\end{figure}
%
\section{Discussion}
In this paper, we investigated the performance of Tyler's M-estimator for subspace recovery, and proved that it recovers the underlying subspace exactly if the percentage of the inliers is larger than a threshold and the data set satisfies a weak assumption on the distribution of data points. We also demonstrated the virtue of this method by simulations and experiments on real data sets.

A future direction is to establish a stronger theoretical guarantee on the robustness of Tyler's M-estimator to noise. Another direction is to extend Tyler's M-estimator for the high-dimensional case. Currently, the iterative update formula  \eqref{eq:IRLS} calculates the inversion of $\bSigma^{(k)}$, which could be prohibitive for large $D$. One may approximate $\bSigma^{(k)}$ by a low-rank matrix in each iteration and reduce the computational cost, but then there is no theoretical guarantee as in Theorem~\ref{thm:exact_recovery}. A method with both reasonable computational complexity for large $D$ and a theoretical guarantee on robust subspace recovery would be very interesting and desired.

\section{Acknowledgement}
The author would like to thank Michael McCoy for reading an earlier version of this manuscript and for helpful comments. The author is grateful to Lek Heng Lim for introducing the book~\cite{bhatia2007positive} and discussions.

\section{Appendix}
\subsection{Proof of Lemma~\ref{lemma:property_objective_function}}\label{sec:property}
\begin{proof}
Geodesic convexity of $F(\bSigma)$ follows from \eqref{eq:convexity} and Lemma~\ref{lemma:convexity_equivalency}. Therefore we only need to prove \eqref{eq:convexity} for geodesic convexity.

We will prove \eqref{eq:convexity} by showing that, if $\bSigma_3\in\PSD(D)$ is the geometric mean of $\bSigma_1,\bSigma_2\in\PSD(D)$, then we have
\begin{equation}\label{eq:det}
\ln(\det(\bSigma_1))+\ln(\det(\bSigma_2))=2\ln(\det(\bSigma_3)),
\end{equation}
and
\begin{equation}\label{eq:convexity2}
\ln(\bx^T\bSigma_1\bx)+\ln(\bx^T\bSigma_2\bx)\geq 2 \ln(\bx^T\bSigma_3\bx).
\end{equation}

We start with the proof of \eqref{eq:det}. Use \eqref{eq:geodesic} with $t=\frac{1}{2}$, we have
\begin{align}
&\bSigma_3\bSigma_1^{-1}\bSigma_3\nonumber\\=&\bSigma_1^{\frac{1}{2}}(\bSigma_1^{-\frac{1}{2}}\bSigma_3\bSigma_1^{-\frac{1}{2}})^\frac{1}{2}\bSigma_1^{\frac{1}{2}}\bSigma_1^{-1}\bSigma_1^{\frac{1}{2}}(\bSigma_1^{-\frac{1}{2}}\bSigma_3\bSigma_1^{-\frac{1}{2}})^\frac{1}{2}\bSigma_1^{\frac{1}{2}}
=\bSigma_2
.\label{eq:deodesic2}
\end{align}
Using~\eqref{eq:deodesic2}, \eqref{eq:det} can be proved as follows:
\begin{align*}
&\det(\bSigma_2)=\det(\bSigma_3\bSigma_1^{-1}\bSigma_3)
=\det(\bSigma_3)\det(\bSigma_1^{-1})\det(\bSigma_3)
\\=&\det(\bSigma_3)^2/\det(\bSigma_1).
\end{align*}
To prove~\eqref{eq:convexity2}, we let the SVD decomposition of $\bSigma_1^{-\frac{1}{2}}\bSigma_2\bSigma_1^{-\frac{1}{2}}=\bU_0\bSigma_0\bU_0^T$ and define $\hat{\bx}=\bU_0\bSigma_1^{\frac{1}{2}}\bx$, then we have $\bx^T\bSigma_1\bx=\hat{\bx}^T\hat{\bx}$, $\bx^T\bSigma_2\bx=\hat{\bx}^T\bSigma_0\hat{\bx}$, and $\bx^T\bSigma_3\bx=\hat{\bx}^T\bSigma_0^{\frac{1}{2}}\hat{\bx}$. Assuming that $\bSigma_0$ is a diagonal matrix with diagonal entries $\sigma_1,\sigma_2,\cdots,\sigma_D$ and $\hat{\bx}=(\hat{x}_1,\hat{x}_2,\cdots,\hat{x}_D)^T$, then \eqref{eq:convexity2} is equivalent to
\[
\sum_{i=1}^{D}\sigma_1\hat{x}_i^2\,\sum_{i=1}^{D}\hat{x}_i^2\geq (\sum_{i=1}^{D}\sigma_1^\frac{1}{2}\hat{x}_i^2)^2,
\]
which can be verified by the Cauchy-Schwartz inequality. Therefore~\eqref{eq:convexity2} is proved.

Finally we investigate the condition such that the equality in \eqref{eq:convexity} holds. By the previous proof of geodesic convexity we know that it holds only when the equality \eqref{eq:convexity2} holds for any $\bx\in\sX$.

By the condition of equality in the Cauchy-Schwartz inequality, we have that the equality in \eqref{eq:convexity} only holds when for any $1\leq i\leq D$ (here $i$ is the index of coordinates) such that $\hat{x}_i\neq 0$, $\sigma_i=c$ for some $c\in\reals$.  When $\bSigma_1\neq c\bSigma_2$, $\sigma_i$ is not the same number for all  $1\leq i\leq D$. Therefore there exists $1\leq i\leq D$ such that $\hat{x}_i=0$. That is, there exists a hyperplane in $\reals^D$ such that $\hat{\bx}$ lies on it. Since $\hat{\bx}$ is a linear transformation of $\bx$, when
\eqref{eq:convexity2} holds for any $\bx\in\sX$, then there exists a hyperplane such that it contains $\bx\in\sX$, which contradicts our assumption that $\mathrm{span}\{\sX\}=\mathbb{R}^D$.
\end{proof}

\subsection{Proof of Lemma~\ref{lemma:singular_infinity}}
\begin{proof}
If Lemma~\ref{lemma:singular_infinity} does not hold, then there exists a sequence $\bSigma_m$ such that it converges some $\tilde{\bSigma}\in\SPSD(D)\setminus\PSD(D)$, and the sequence $F(\bSigma_m)$ is bounded. WLOG we assume that $\lambda_j(\bSigma_{m_i})$ and $\bv_j(\bSigma_{m_i})$ also converge for any $1\leq j\leq p$, where $\lambda_j(\bSigma)$ and $\bv_j(\bSigma)$ are the
$j$-th eigenvalue and eigenvector of $\bSigma$. This can be assumed since any sequence has a subsequence satisfying this property (eigenvectors and eigenvalues of $\bSigma_m$ lie in a compact space).

We prove \eqref{eq:convergence_singular} by induction on the ambient dimension $D$. When $D$=2, we have $\dim(\ker(\tilde{\bSigma}))=1$, and
\begin{align}\label{eq:induction_1}
F(\bSigma_m)\geq&
\frac{1}{N}\sum_{\bx\in\sX\setminus\ker(\tilde{\bSigma})}
\Big(\log(\lambda_2(\bSigma_m))+2\log(\bx^T \bv_2(\bSigma_m))\Big)\\+&\nonumber\frac{1}{2}\log(\lambda_2(\bSigma_m))+
\frac{1}{2}\log(\lambda_1(\bSigma_m))
.
\end{align}
When $\bx\notin\ker(\tilde{\bSigma})$, we have $\liminf_{m\rightarrow\infty}\bx^T \bv_2(\bSigma_m)>0$, therefore the term $\log(\bx^T \bv_2(\bSigma_m))$ is bounded from below. Applying the assumption that $\lambda_1(\bSigma_m)$ are bounded from below, $\frac{1}{2}\log(\lambda_1(\bSigma_m))$ is also bounded from below. Applying the assumption $\frac{|\sX\setminus\ker(\tilde{\bSigma})|}{N}>\frac{1}{2}$ and  $\lim_{m\rightarrow\infty}\lambda_2(\bSigma_m)=0$, the RHS of \eqref{eq:induction_1} converges to $+\infty$, which is a contradiction to the assumption that $F(\bSigma_m)$ is bounded, and therefore \eqref{eq:convergence_singular} is proved.

If \eqref{eq:convergence_singular} holds for the case $\dim(\bx)<D_0$, then we will prove \eqref{eq:convergence_singular} for $\dim(\bx)=D_0$.
By the assumption on the convergence of eigenvectors and eigenvalues of $\bSigma^{(k)}$, to prove \eqref{eq:convergence_singular} it is equivalent to prove that \begin{equation}\text{$F'(\bSigma'_m)\rightarrow\infty$ as $m\rightarrow\infty$,}\label{eq:induction}\end{equation} where $\bSigma'_m=\bP_{\tilde{L}}^T\bSigma_m\bP_{\tilde{L}}$, ${\tilde{L}}=\ker(\tilde{\bSigma})$, $d_0=\dim(\tilde{L})$ and $F': \PSD(d_0)\rightarrow\reals$ is defined by
\[
F'(\bSigma)=\frac{1}{N}\sum_{\bx\in\sX}
\log((\bP_{\tilde{L}}^T\bx)^T\bSigma^{-1}\bP_{\tilde{L}}^T\bx)+\frac{1}{D_0}\log\det(\bSigma).
\]

An important observation is that $\lim_{m\rightarrow\infty}\trace({\bSigma}'_m)=0$. Combine it with $\frac{|\sX\setminus\tilde{L}|}{N}>\frac{d_0}{D_0}$, we have
\begin{align}
\lim_{m\rightarrow\infty} F'({\bSigma}'_m)-F'(\tilde{\bSigma}'_m)= (\frac{d_0}{D_0}-\frac{|\sX\setminus\tilde{L}|}{N})\lim_{m\rightarrow\infty}\log\tr({\bSigma}'_m)
=\infty.\label{eq:induction1}\end{align}

When $\tilde{\bSigma}'_m$ converges to a nonsingular matrix $\tilde{\bSigma}'$,
\begin{equation}\label{eq:induction2}
\lim_{m\rightarrow\infty} F'(\tilde{\bSigma}'_m)=F'(\tilde{\bSigma}')=C\end{equation}
for some constant $C$, and when $\tilde{\bSigma}'_m$ converges to a singular matrix, by induction
\begin{equation}\label{eq:induction3}
\lim_{m\rightarrow\infty} F'(\tilde{\bSigma}'_m)=\infty.
\end{equation}
Combining \eqref{eq:induction1}, \eqref{eq:induction2} and \eqref{eq:induction3}, \eqref{eq:induction} is proved and therefore Lemma~\ref{lemma:singular_infinity} is proved by induction.

\end{proof}

\subsection{Proof of Theorem~\ref{thm:stability1}}\label{sec:thm_stability1}
Let $\bT=\hat{\bSigma}^{-1/2}$ and $\tilde{\bx}_i=\bT\bx_i/\|\bT\bx_i\|$, then by diffrentiating the objective function of Tyler's M-estimator, we have
\begin{equation}\label{eq:derivative}
\hat{\bSigma}\sum_{i=1}^N\frac{\bx_i^T\bx_i}{\bx_i^T\hat{\bSigma}^{-1}\bx_i}=c\bI,\,\,\,\text{for some $c\in\reals$},
\end{equation}
which means
\begin{equation}\label{eq:by0}
\sum_{i=1}^N \tilde{\bx}_i\tilde{\bx}_i^T=c\bI,\,\,\,\text{for some $c\in\reals$}.
\end{equation}
By compareing the trace of LHS and RHS of \eqref{eq:by0} we have $c=\frac{N}{D}$ and
\begin{equation}\label{eq:by}
\sum_{i=1}^N \tilde{\bx}_i\tilde{\bx}_i^T=\frac{N}{D}\bI.
\end{equation}
If $\hat{\bSigma}$ is singular then we can proceed with the following proof by treating the range of $\hat{\bSigma}$ as the ambient space, and instead of $\sX$,  considering the subset of $\{\bx_i\}_{i=1}^N$ that lie in the range.

In the following proof, we use $\bT(L)$ to denote the image of the subspace $L$ after the transformation $\bT$, which is a subspace with the same dimensionality of $L$.

Since \begin{equation}\label{eq:bT}
\dist(\bT\bx,\bT({L}_*))\leq \dist(\bT\bx, \bT (\bP_{L_*}\bx) )\leq \|\bT\|\dist(\bx,L_*),
\end{equation}
\[
\dist(\bx,L_*)\leq \dist(\bT^{-1}(\bT\bx), \bT^{-1}\, \bT (\bP_{L_*}\bx) ) \leq \|\bT^{-1}\|\dist(\bT\bx,\bT({L}_*)),
\]
and the above inequalities holds when $L_*$ and $\bT({L}_*)$ are replaced by $L_*^\perp$ and $\bT({L}_*)^\perp$, for $1\leq i\leq N_1$ we have
\[
\frac{\dist(\tilde{\bx}_i,\bT(L_*))}{\dist(\tilde{\bx}_i,\bT(L_*)^\perp)}\leq \|\bT\|\|\bT^{-1}\| \frac{\dist(\tilde{\bx}_i,L_*)}{\dist(\tilde{\bx}_i,L_*^\perp)}\leq \|\bT\|\|\bT^{-1}\|\frac{\eps}{\sqrt{1-\eps^2}}=\kappa(\bT)\frac{\eps}{\sqrt{1-\eps^2}},
\]
where $\kappa(\bT)= \|\bT\|\|\bT^{-1}\|$ is the conditional number of $\bT$.
Therefore, $\|\bP_{\bT(L_*)}\tilde{\bx}_i\|\geq \frac{\sqrt{1-\eps^2}}{\sqrt{1+(\kappa(\bT)^2-1)\eps^2}}$ for $1\leq i\leq N_1$ and applying \eqref{eq:by} we have
\begin{align*}
\frac{N}{D}d=\tr\Big(\bP_{\bT(L_*)}(\sum_{i=1}^N \tilde{\bx}_i\tilde{\bx}_i^T)\bP_{\bT(L_*)}^T\Big)
\geq \sum_{i=1}^{N_1} \|\bP_{\bT(L_*)}\tilde{\bx}_i\|^2
\geq N_1\frac{{1-\eps^2}}{{1+(\kappa(\bT)^2-1)\eps^2}}.
\end{align*}
Therefore,
\begin{equation}\label{eq:conditionalnumber}
\kappa(\hat{\Sigma})=\kappa(\bT)^2\geq (\alpha-1)\,\frac{1-\eps^2}{\eps^2}.
 \end{equation} 

The rest of the proof of Theorem~\ref{thm:stability1} is based on the following lemmas, and their proof are deferred.
\begin{lemma}\label{lemma:gap0}If  \begin{equation}\label{eq:eigengap0}\frac{\lambda_m(\hat{\Sigma})}{\lambda_{m+1}(\hat{\Sigma})} > \beta^2\end{equation} for some $1\leq m\leq D$, then for any $\eta>0$, there would be at least $\frac{m\eta^2\beta^2+m-D}{D\eta^2\beta^2}N$  points satisfying
\begin{equation}\label{eq:eigengap}
\frac{\dist(\bx_i,L_m)}{\dist(\bx_i,L_m^\perp)}\leq \eta,
\end{equation}
where $L_m$ is the $m$-dimensional subspace spanned by the top $m$ eigenvectors of $\hat{\Sigma}$.
\end{lemma}
\begin{lemma}\label{lemma:gap1}
Let
 \begin{equation}\label{eq:beta}
\beta_0=\max\big(2,4 C_1^2, \frac{D-d}{5}, {25\,C_2^2(C_1+2)^2},{c_1}^{-1},\eta_0^{-2} \big),
\end{equation}
then for any $1\leq m\leq d-1$,  $ \frac{\lambda_m(\hat{\Sigma})}{\lambda_{m+1}(\hat{\Sigma})} < \beta_0^2.
$

Besides, if there exists some $m> d$ such that
$
 \frac{\lambda_m(\hat{\Sigma})}{\lambda_{m+1}(\hat{\Sigma})} > \beta_0^2,$
then
 \[
 \frac{\lambda_1(\hat{\Sigma})}{\lambda_{m}(\hat{\Sigma})}> \frac{4-7\eps^2}{4\eps^2}\cdot \frac{m-d}{10\,d}.
 \]
\end{lemma}

With Lemma~\ref{lemma:gap0} and Lemma~\ref{lemma:gap1}, we are ready to prove Theorem~\ref{thm:stability1}. Assuming that $m_0$ is the smallest number such that $m_0>d$ and $ \frac{\lambda_{m_0}(\hat{\Sigma})}{\lambda_{{m_0}+1}(\hat{\Sigma})} > \beta_0^2$. If such ${m_0}$ exists, then by Lemma~\ref{lemma:gap1}, $\frac{\lambda_1(\hat{\Sigma})}{\lambda_{m_0}(\hat{\Sigma})} > \frac{4-7\eps^2}{4\eps^2}\cdot \frac{m_0-d}{10\,d}$. By the definition of $m_0$ and Lemma~\ref{lemma:gap1},
$\frac{\lambda_{i}(\hat{\Sigma})}{\lambda_{i+1}(\hat{\Sigma})} <\beta_0^2$ for all $1\leq i\leq m_0-1$ except for $i=d$. Therefore,
\[
\frac{\lambda_{d}(\hat{\Sigma})}{\lambda_{d+1}(\hat{\Sigma})}
\geq  \frac{4-7\eps^2}{4\eps^2}\cdot \frac{m_0-d}{10\,d\beta_0^{2D-2}}.
\]

If there does not exist $m_0>d$ such that $ \frac{\lambda_{m_0}(\hat{\Sigma})}{\lambda_{m_0+1}(\hat{\Sigma})} > \beta_0^2$, then by \eqref{eq:conditionalnumber} we have
\[
\frac{\lambda_{d}(\hat{\Sigma})}{\lambda_{d+1}(\hat{\Sigma})}
\geq\,\frac{1-\eps^2}{\eps^2}\cdot \frac{\alpha-1}{\beta_0^{2D-2}}.
\]
Combining these two cases with the assumption that $\eps<1/2$ we have
\begin{equation}\label{eq:C0}
\frac{\lambda_{d}(\hat{\Sigma})}{\lambda_{d+1}(\hat{\Sigma})}\geq
C_0/\eps^2,  \text{where $C_0 = \min\Big( \frac{m-d}{20\,d\beta_0^{2D-2}}\,\,\,,\,\, \frac{\alpha-1}{2\,\beta_0^{2D-2}}\Big)$}
\end{equation}
By Lemma~\ref{lemma:gap0}, there are at least $(\frac{d}{D}-\frac{(D-m)\eps}{D\sqrt{C_0}})N$ points such that $\dist(\bx_i,L_d)\leq \frac{\dist(\bx_i,L_d)}{\dist(\bx_i,L_d^\perp)}\leq \frac{\eps}{\sqrt{C_0}}$. Combining it with the Assumption 2 and the assumptions that $\frac{(D-m)\eps}{D\sqrt{C_0}}<c_1$ and $\frac{\eps}{\sqrt{C_0}}<\eta_0$, we have $\|\bP_{L_d}-\bP_{L_*}\|<\frac{\eps\,C_1}{\sqrt{C_0}}$.

\subsubsection{The proof of Lemma~\ref{lemma:gap0}}
\begin{proof}By the definition of $L_m$ and the eigengap in \eqref{eq:eigengap0}, it is easy to verify that
\[
\beta\frac{\dist(\bx,L_m)}{\dist(\bx,L_m^\perp)}\leq \frac{\dist(\tilde{\bx},\bT(L_m))}{\dist(\tilde{\bx},\bT(L_m)^\perp)}.
\]
Therefore, for any $\bx_i$ such that $\frac{\dist(\bx_i,L_m)}{\dist(\bx_i,L_m^\perp)}>\eta$, we have $\frac{\dist(\tilde{\bx}_i,\bT(L_m))}{\dist(\tilde{\bx}_i,\bT(L_m)^\perp)}>\eta\beta$ and \begin{equation}\|\bP_{\bT(L_m)^\perp}\tilde{\bx}_i\|^2>\frac{\eta^2\beta^2}{\eta^2\beta^2+1}.\label{eq:projection1}\end{equation} Combing \eqref{eq:projection1} with the fact that $\sum_{i=1}^N\|\bP_{\bT(L_m)^\perp}\tilde{\bx}_i\|^2=\frac{D-m}{D}N$, there are at most $\frac{(D-m)}{D}N\cdot\frac{(\eta^2\beta^2+1)}{D\eta^2\beta^2}$ points that violate \eqref{eq:eigengap} and therefore, there are at least $\frac{m\eta^2\beta^2+m-D}{D\eta^2\beta^2}N$ points satisfying \eqref{eq:eigengap}.
\end{proof}
\subsubsection{The proof of Lemma~\ref{lemma:gap1}}
\begin{proof} If $
 \frac{\lambda_m(\hat{\Sigma})}{\lambda_{m+1}(\hat{\Sigma})} > \beta_0^2,$ then by applying Lemma~\ref{lemma:gap0} with $\beta=\beta_0$ and $\eta=\beta_0^{-\frac{1}{2}}$,
there are at least $(\frac{m}{D}-\frac{D-m}{D\beta_0})N$ points such that \eqref{eq:eigengap} is satisfied for $L_m$. Applying the Assumption 2 with
$ \frac{D-m}{D\beta_0}< c_1$ and $\beta_0^{-\frac{1}{2}}<\eta_0$,
 we have $m\geq d$, and
 \[
 \|\bP_{L_*}-\bP_{P_{L_m}L_*}\|<C_1\beta_0^{-\frac{1}{2}}.
 \]

 Now we only need to consider the case $m>d$. Divide the set of indices $\{1, 2,\cdots, N\}$ into three subsets
 $\sI_1=\{1, 2, \cdots, N_1\}$, $\sI_2=\{i: N_1+1 \leq i\leq N, \frac{\dist(\bx_i,L_m)}{\dist(\bx_i,L_m^\perp)}>\beta_0^{-\frac{1}{2}}
\}$, and $\sI_3=\{i: N_1+1\leq i\leq N, \frac{\dist(\bx_i,L_m)}{\dist(\bx_i,L_m^\perp)}\leq \beta_0^{-\frac{1}{2}}
\}$.
Applying Lemma~\ref{lemma:gap0} we have
\begin{equation} \text{$|\sI_2|\leq\frac{(D-m)(\beta_0+1)}{D\beta_0}N$.}\label{eq:est_Y}\end{equation}

Let the subspace $L_m'$ defined by $L_m' = (L_m\ominus P_{L_m}L_*) \oplus L_*$, where $L_1\ominus L_2=L_1\cap L_2^\perp$ and $L_1\oplus L_2= \{\bx_1+\bx_2: \bx_1\in L_1, \bx_2\in L_2\}$. Since $|\dist(\bx_i,L_m')-\dist(\bx_i,L_m)|\leq \|\bP_{L_m}-\bP_{L_m'}\|= \|\bP_{L_*}-\bP_{P_{L_m}L_*}\|<C_1/\sqrt{\beta_0}$, for any $i\in\sI_3$ we have $\dist(\bx_i,L_m)<\frac{1}{\sqrt{\beta_0-1}}$ and therefore $\dist(\bx_i,L_m')<C_1/\sqrt{\beta_0}+ \frac{1}{\sqrt{\beta_0-1}}<(C_1+2)/\sqrt{\beta_0}$, where the last step uses the assumption that $\beta_0>2$. Applying Assumption 3 to $L_m'$ with $\eta=(C_1+2)/\sqrt{\beta_0}$, we have
\begin{equation} \text{$|\sI_3|\leq \max(C_2(C_1+2) N/\sqrt{\beta_0}, m-d )$.}\label{eq:est_Y1}\end{equation}

Now let us consider $\sum_{i=1}^N \|P_{\bT(P_{L_m}L_*)}\tilde{\bx}_i\|^2$ and $\sum_{i=1}^N \|P_{\bT(L_m)\ominus \bT(P_{L_m}L_*)}\tilde{\bx}_i\|^2$. When $i\in\sI_2$, applying \eqref{eq:projection1} we have
\begin{equation}\label{eq:est_sY2}
\|P_{\bT(L_m)}\tilde{\bx}_i\|^2\leq \frac{1}{\beta_0+1}.
\end{equation}
Combining \eqref{eq:est_Y}, \eqref{eq:est_Y1} and \eqref{eq:est_sY2} we have
\begin{equation}\label{eq:est_Y4}
\sum_{i\in\sI_2}\|P_{\bT(L_m)}\tilde{\bx}_i\|^2+\sum_{i\in\sI_3}\|P_{\bT(L_m)}\tilde{\bx}_i\|^2
\\\leq M, \,\,\,\text{where $M=\frac{(D-m)}{D\beta_0}N + \max(C_2(C_1+2) N/\sqrt{\beta_0}, m-d )$}.
\end{equation}
Combining \eqref{eq:est_Y4} with $\sum_{i=1}^N \|P_{\bT(L_m)}\tilde{\bx}_i\|^2 = \frac{m}{D}N$, we have \begin{align*}&\sum_{i\in\sI_1}\|P_{\bT(L_m)}\tilde{\bx}_i\|^2 =
\sum_{1\leq i\leq N}\|P_{\bT(L_m)}\tilde{\bx}_i\|^2-\sum_{i\in\sI_2}\|P_{\bT(L_m)}\tilde{\bx}_i\|^2-\sum_{i\in\sI_3}\|P_{\bT(L_m)}\tilde{\bx}_i\|^2
\geq \frac{m}{D}N - M.\end{align*}
Combining it with the estimation from Lemma~\ref{lemma:est_Y1} that
\[
\frac{\sum_{i\in\sI_1}\|P_{\bT(L_m)\ominus \bT(P_{L_m}L_*)}\tilde{\bx}_i\|^2}{\sum_{i\in\sI_1}\|P_{\bT(P_{L_m}L_*)}\tilde{\bx}_i\|^2}
\leq \frac{\kappa_m\eps^2}{{((1-\eps^2)(1-C_1^2/\beta_0)-\eps^2)}},
\]
where  $\kappa_m=\frac{\lambda_1(\hat{\bSigma})}{\lambda_m(\hat{\bSigma})}$, we have
\begin{equation}\label{eq:M1}
\sum_{i\in\sI_1}\|P_{\bT(L_m)\ominus \bT(P_{L_m}L_*)}\tilde{\bx}_i\|^2\geq  \frac{\kappa_m\eps^2( \frac{m}{D}N - M)}{{((1-\eps^2)(1-C_1^2/\beta_0)-\eps^2)}+\kappa_m\eps^2}.
\end{equation}
Applying \eqref{eq:M1} and \eqref{eq:est_Y4}, we have
\[
\frac{m-d}{D}N=\sum_{i\in\sI_1\cup\sI_2\cup\sI_3}\|P_{\bT(L_m)\ominus \bT(P_{L_m}L_*)}\tilde{\bx}_i\|^2
\leq\frac{\kappa_m\eps^2( \frac{m}{D}N - M)}{{((1-\eps^2)(1-C_1^2/\beta_0)-\eps^2)}+\kappa_m\eps^2} + M.
\]
Therefore,
\begin{equation}\label{eq:kappam}
\kappa_m\geq \frac{(1-\eps^2)(1-C_1^2/\beta_0)-\eps^2}{\eps^2}\,\cdot\,
\frac{(m-d)N-DM}{d N}.
\end{equation}
By the definition of $\beta_0$ and the assumption $N>2D$, we have $C_1^2/\beta_0>1/4$, $(m-d)D<\frac{1}{2}(m-d)N$, $D\,\frac{D-m}{\beta_0} < \frac{1}{5}(m-d)N$ and $D\, C_2(C_1+2)/\sqrt{\beta_0}<\frac{1}{5}(m-d)$. Therefore,
\[
\frac{(1-\eps^2)(1-C_1^2/\beta_0)-\eps^2}{\eps^2}\geq \frac{4-7\eps^2}{4\eps^2}
\]
and
\[
(m-d)N-DM  > \frac{m-d}{10}N.
\]
The proof of Lemma~\ref{lemma:gap1} follows from these estimations and \eqref{eq:kappam}.
\end{proof}

\begin{lemma}\label{lemma:est_Y1}
For $1\leq i\leq N_1$, we have
\begin{equation}\label{eq:lemma_est_Y11}\frac{\|P_{L_m\ominus P_{L_m}L_*}\bx_i\|}{\|P_{P_{L_m}L_*}\bx_i\|}
\leq \frac{\eps}{\sqrt{(1-\eps^2)(1-C_1^2/\beta_0)-\eps^2}}.
\end{equation}\begin{equation}\label{eq:lemma_est_Y12}
\frac{\|P_{\bT(L_m)\ominus \bT(P_{L_m}L_*)}\tilde{\bx}_i\|}{\|P_{\bT(P_{L_m}L_*)}\tilde{\bx}_i\|}
\leq \frac{\eps}{\sqrt{\kappa_m((1-\eps^2)(1-C_1^2/\beta_0)-\eps^2)}},
\end{equation}
where $\kappa_m=\frac{\lambda_1(\hat{\bSigma})}{\lambda_m(\hat{\bSigma})}$.
\end{lemma}
\begin{proof}
Since
\[
\dist(P_{L_*}\bx_i, P_{P_{L_m}L_*}(P_{L_*}\bx_i))=\dist(P_{L_*}\bx_i, P_{L_m}L_*)\leq \|\bP_{L_*}-\bP_{P_{L_m}L_*}\|\|P_{L_*}\bx_i\|\leq C_1/\sqrt{\beta_0}\cdot 1 = C_1/\sqrt{\beta_0}
\]
and
\[
\dist(\bx_i,P_{L_*}\bx_i)\leq \eps,
\]
we have
\[
\|P_{L_m}\bx_i\| = \sqrt{1-\dist(\bx_i,L_m)^2}
\geq \sqrt{1-\dist(\bx_i,P_{P_{L_m}L_*}(P_{L_*}\bx_i))}\geq \sqrt{1-(C_1/\sqrt{\beta_0}+\eps)^2}.
\]
Combining it with \[
\|P_{L_m\ominus P_{L_m}L_*}\bx_i\|=\dist(P_{L_m}\bx_i, P_{L_m}L_*)
\leq \dist(\bx_i,L_*)\leq \eps,
\] and
\[
\|P_{P_{L_m}L_*}\bx_i\|^2=\|P_{L_m}\bx_i\|^2-\|P_{L_m\ominus P_{L_m}L_*}\bx_i\|^2,
\]
\eqref{eq:lemma_est_Y11} is proved.
Combining \eqref{eq:lemma_est_Y11} with the same argument in \eqref{eq:bT} ($\kappa$ in \eqref{eq:bT} is replaced in $\kappa_m$ since the transformation $\bT$ in \eqref{eq:lemma_est_Y12} in restricted on the subspace $L_m$), \eqref{eq:lemma_est_Y12} is proved.
\end{proof}

\bibliographystyle{abbrv}
\bibliography{bib-rrp}
\end{document}